\documentclass{article}
\usepackage{ijcai22}

\usepackage{times}
\usepackage{soul}
\usepackage[utf8]{inputenc} 
\usepackage[hidelinks]{hyperref}
\usepackage{url}            
\usepackage{booktabs}       
\usepackage{amsfonts}       
\usepackage{nicefrac}       
\usepackage{microtype}      
\usepackage{xcolor}         
\usepackage{comment}
\usepackage{graphicx}
\usepackage[small]{caption}
\usepackage{subcaption}
\usepackage{parskip}
\usepackage{algorithm}
\usepackage{algpseudocode}

\usepackage{amsmath,amsfonts,amssymb}
\usepackage{amsthm}
\usepackage{booktabs}
\usepackage{multirow}
\usepackage{graphicx}
\usepackage{cleveref}
\usepackage{highlight}
\usepackage{amsthm}
\usepackage{makecell}
\usepackage[english]{babel}
\usepackage{blindtext}
\usepackage{float}
\usepackage{makecell}
\theoremstyle{definition}

\algnewcommand\algorithmicswitch{\textbf{switch}}
\algnewcommand\algorithmiccase{\textbf{case}}
\algnewcommand\algorithmicassert{\texttt{assert}}
\algnewcommand\Assert[1]{\State \algorithmicassert(#1)}%
\algdef{SE}[SWITCH]{Switch}{EndSwitch}[1]{\algorithmicswitch\ #1\ \algorithmicdo}{\algorithmicend\ \algorithmicswitch}%
\algdef{SE}[CASE]{Case}{EndCase}[1]{\algorithmiccase\ #1}{\algorithmicend\ \algorithmiccase}%
\algtext*{EndSwitch}%
\algtext*{EndCase}%

\title{Visual Transformer Meets CutMix for Improved Accuracy, \\Communication Efficiency, and Data Privacy in Split Learning}
\author{
Sihun Baek$^1$
\and
Jihong Park$^2$\and
Praneeth Vepakomma$^3$\\\
Ramesh Raskar$^3$ \and
Mehdi Bennis$^4$ \and and
Seong-Lyun Kim$^1$
\affiliations
$^1$Yonsei University, Seoul, South Korea\\
$^2$Deakin University, Geelong, Victoria, Australia\\
$^3$MIT Media Lab, Cambridge, Massachusetts, USA\\
$^4$University of Oulu, Oulu, Finland\\
\emails
shbaek@ramo.yonsei.ac.kr, 
jihong.park@deakin.edu.au, vepakom@mit.edu\\
raskar@media.mit.edu, mehdi.bennis@oulu.fi, slkim@yonsei.ac.kr
}

\begin{document}
\maketitle
\begin{abstract}
This article seeks for a distributed learning solution for the \textit{visual transformer (ViT)} architectures. Compared to convolutional neural network (CNN) architectures, ViTs often have larger model sizes, and are computationally expensive, making federated learning (FL) ill-suited. Split learning (SL) can detour this problem by splitting a model and communicating the hidden representations at the split-layer, also known as smashed data. Notwithstanding, the smashed data of ViT are as large as and as similar as the input data, negating the communication efficiency of SL while violating data privacy. To resolve these issues, we propose a new form of \textit{CutSmashed data} by randomly punching and compressing the original smashed data. Leveraging this, we develop a novel SL framework for ViT, coined \textit{CutMixSL}, communicating CutSmashed data. CutMixSL not only reduces communication costs and privacy leakage, but also inherently involves the CutMix data augmentation, improving accuracy and scalability. Simulations corroborate that CutMixSL outperforms baselines such as parallelized SL and SplitFed that integrates FL with SL.
\end{abstract}
\section{Introduction}
Transformer architectures have revolutionized various application domains in deep learning, ranging from natural language processing (NLP) \cite{vaswani2017attention} to speech recognition \cite{karita2019comparative}. Fueled by this success, recently there has been another paradigm shift in computer vision (CV) where the \emph{visual transformer (ViT)} architecture has broken the performance record set by the \emph{de facto} standard convolutional neural network (CNN) architectures \cite{dosovitskiy2020image}. The core idea of ViT is to divide each input image sample into multiple \emph{patches} as in the tokens in NLP, and to process the patches in parallel using the \emph{attention} mechanism of Transformer. This process is in contrast to that of CNN which processes only neighboring pixels in sequence using the convolutional operations. 

While the existing studies focus mostly on centralized ViT operations \cite{han2022survey}, in this article we delve into the problem of distributed learning with ViTs that are often computationally expensive, and require more training samples than CNNs \cite{khan2021transformers,han2022survey}. In the recent literature of distributed learning, federated learning (FL) is one promising solution that enables training a global model across edge devices such as phones, cameras, and the Internet of Things (IoT) devices \cite{li2020federated}. The key idea is periodically averaging model parameters across edge devices or clients through a parameter server, without directly exchanging private data. However, model averaging requires every client to store and communicate the entire model, so is ill-suited for ViT due to its large model size. 

Alternatively, each client can store only a fraction of the entire model, and offload the remaining segment onto the server under a model-split architecture \cite{gupta2018distributed}. \emph{Split learning (SL)} follows this model-split parallelism, under which in the forward propagation (FP), each client uploads the hidden layer activations at the split-layer, also known as \emph{smashed data}, followed by downloading gradients in the back propagation (BP) \cite{vepakomma2018split}. Unfortunately, as opposed to CNN's smashed data distorted by convolution operations, ViT's smashed data wihtout convolution look similar to their input data \cite{yuan2021tokens}, which may leak private information on raw data to the server. Furthermore, ViT often lacks pooling layers, so the smashed data sizes are as large as the input data, negating the communication efficiency of SL.

\begin{figure*}[t]
\centering
\begin{subfigure}{0.24\textwidth}
\centering
\includegraphics[width=\textwidth]{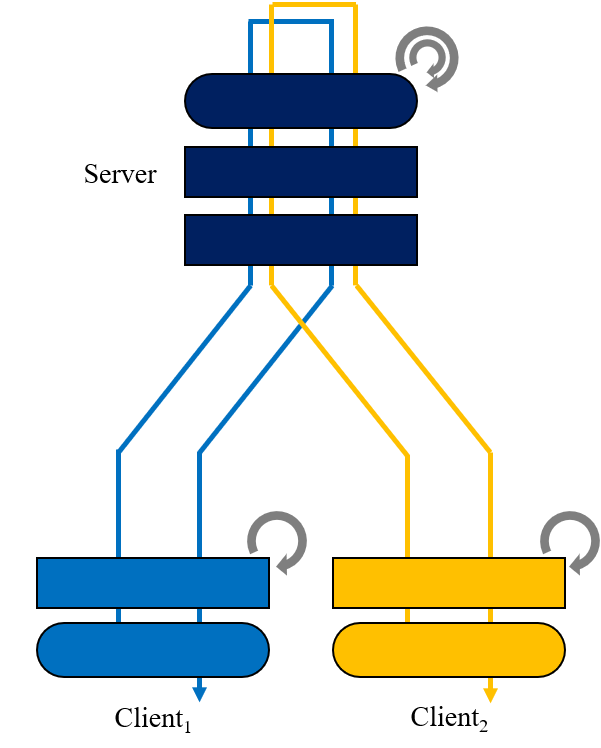}
\caption{Baseline 1: Parllel SL.}
\label{Paralllel SL.}
\end{subfigure}
\hfill
\begin{subfigure}{0.24\textwidth}
\centering
\includegraphics[ width=\textwidth]{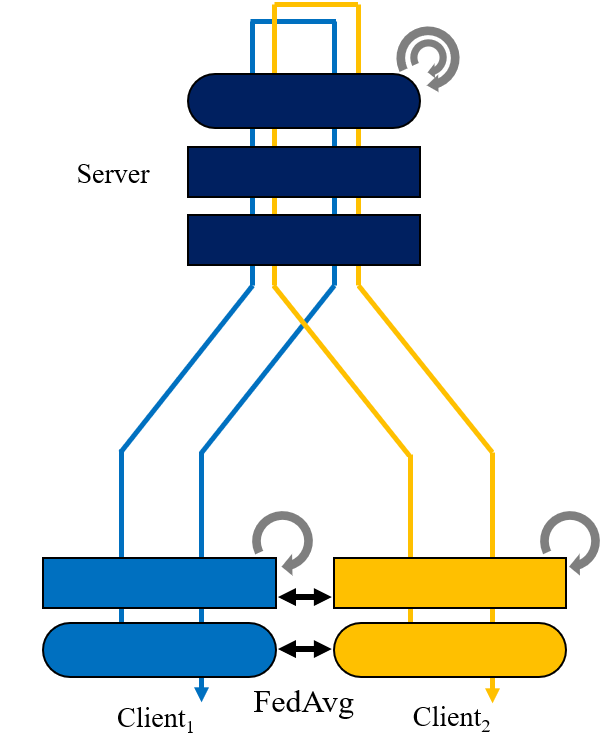}
\caption{Baseline 2: SFL.}
\label{federated learning}
\end{subfigure}
\hfill
\begin{subfigure}{0.48\textwidth}
\centering
\includegraphics[width=\textwidth]{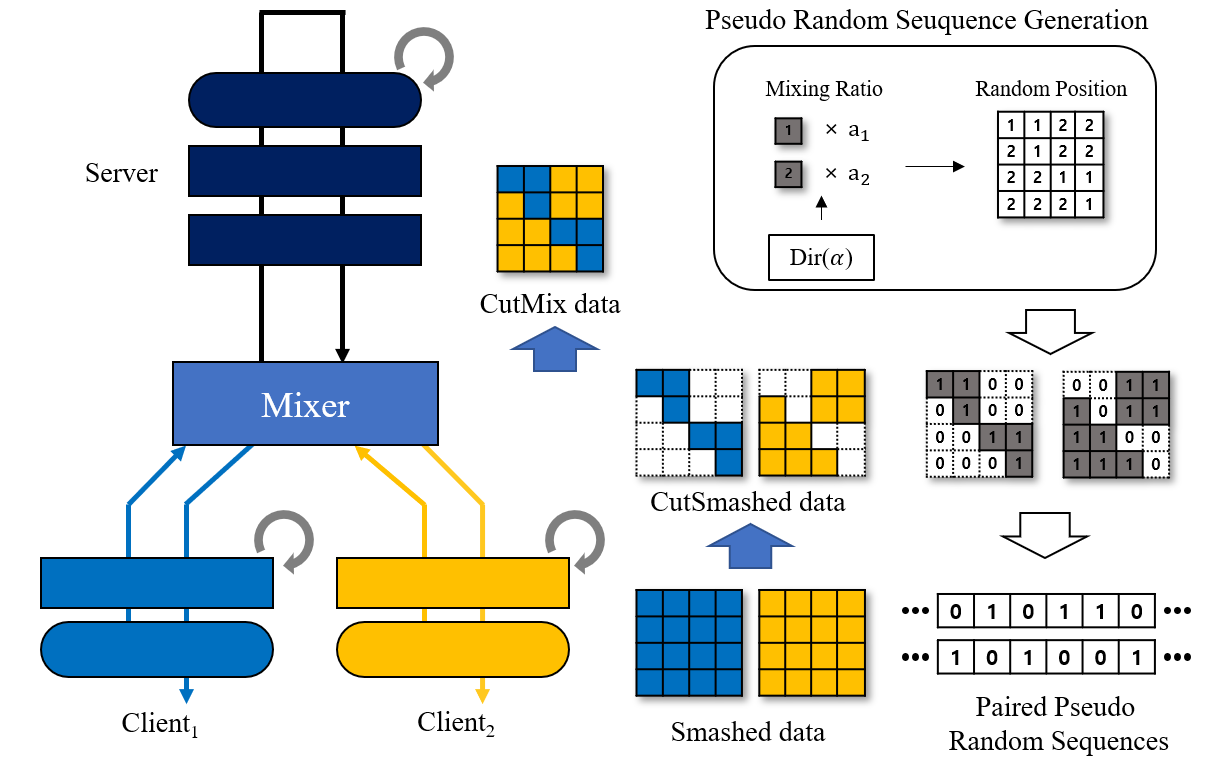}
\caption{Proposed: CutMixSL.}
\label{fig:vit}
\end{subfigure}
\hfill
\caption{Schematic illustrations of: (a) parallel split learning (SL), (b) SplitFed (SFL), and (c) CutMixSL.}
\end{figure*}

To resolve the aforementioned issues, inspired from the \emph{CutMix} data augmentation technique \cite{yun2019cutmix}, we propose a new type of \emph{CutSmashed data}, and thereby develop a novel split learning framework for ViTs, coined \emph{CutMixSL}. In CutMixSL, each client constructs the CutSmashed data by randomly masking the patches of the original smashed data. For instance, Fig. 1 illustrates two clients locally constructing the CutSmashed data by applying the mutually exclusive masks $010110$ and $101001$ to their original smashed data, respectively. Here, to guarantee the mutual exclusiveness and the subsequent operations, a common pseudo random sequence generator is shared by all clients and the server, without sharing raw data. Then, the mixer at the server adds these uploaded CutSmashed data, resulting in \emph{CutMix data} (i.e., with the mask $111111$) that FP in the server, followed by BP from the server to clients. 

The key benefits of CutMixSL are summarized as follows.
\begin{itemize}
\item First, the clients in CutMixSL can only upload non-zero masked CutSmashed data, \emph{reducing privacy leakage}.

\item Second, CutMix data plays its original role as data augmentation complementing ViT's lack of useful inductive bias as opposed to CNNs, thereby \emph{improving accuracy}.

\item Lastly, CutMixSL is free from the standard paralell SL's imbalance problem between the server-side and client-side model updates \cite{oh2022locfedmix,pal2021server}, \emph{achieving scalability} in temrs of the accuracy increasing with the number of clients.

\item Simulations under the CIFAR-10 classification task show that CutMixSL reduces the privacy leakage (measured by reconstruction mean-squared errors) by around 8$\times$, decreases uplink communication payload sizes by $20-50\%$, and improves accuracy by up to $18.5$\% while achieving scalability up to (at least) $10$ clients.

\end{itemize}


\section{Related works}
\paragraph{FL, SL, and Hybrids} FL and SL are two popular distributed machine learning techniques preserving data privacy by avoiding raw data exchanges \cite{konevcny2015federated,gupta2018distributed}. These methods have their own advantages and disadvantages. FL is scalable in the sense that the accuracy increases with the number of clients \cite{konevcny2015federated}. However, FL requires exchanging and averaging model parameters, so cannot handle large models under limited memory, computing, and communication resources \cite{konevcny2016federated}. On the other hand, SL is applicable to large models by means of its spliting the entire model into segments, e.g.,  two segments stored at the server and at each client, respectively \cite{gupta2018distributed,vepakomma2018split}.
To enjoy the advantages of both methods, SplitFed (SFL) integrates FL into SL by averaging only clients' model segments after BP \cite{DBLP:journals/corr/abs-2004-12088,gao2021evaluation}, at the cost of increasing communication overhead. Alternatively, a similar effect can be achieved by averaging or mixing the hidden activations during FP \cite{xiao2021mixing} and \cite{oh2022locfedmix}. Our proposed method focuses primarily on the activation mixing, and the segmentation averaging is supplementary. 

\paragraph{Transformer and ViT}
Transformer is the first architecture driven entirely by the attention mechanism \cite{vaswani2017attention}, making remarkable success in NLP as evidenced by BERT \cite{devlin2018bert} and GPT-3 \cite{brown2020language}. Motivated by this, Transformer has expended its applications to various domains, among which ViT is the recent success made in CV \cite{DBLP:journals/corr/abs-2010-11929}. In CV, there are other architectures that partly utilize the concepts of Transformer \cite{carion2020end,zheng2021rethinking}. As opposed to them, ViT is a Trasformer-native architecture, which is viable by splitting and converting an input image into the input sequence of image patches as analogous to the input sequence of language tokens in NLP. Departing from standalone operations, distributed Transformer has been studied in the context of FL and SL for various applications. Transformer with FL has been investigated for a Text-to-Speech (TTS) task in \cite{hong2021federated}. Transformer with SL has been developed for a COVID-19 diagnosis in \cite{park2021federated}. ViT with FL has also been studied under heterogeneous data in \cite{qu2021rethinking}. Inspired from these preceding studies, for the first time, we investigate ViT with SL as well as its parallel, communication-efficient, and scalable architecture.

\section{Token-based Split Learning for ViT}
In this paper, a novel split learning based on patches where transformers operate is proposed to solve the issue of heavy communication payload and data privacy leakage problem. We describe components of the proposed learning algorithm in an order of training sequence. The overall procedure of the proposed is visualized as \Cref{fig:vit}.

\subsection{CutSmashed data for Communication Efficiency and Privacy Enhancement}
\label{partialuploading}


To increase a communication efficiency, and to resolve the data privacy leakage problem, we pay attention to the way how transformers address data. Transformers divide data into multiple separate tokens and process the entire range of sequence in parallel. In this sense, we instinctively expect that cutting off a certain amount of patches of smashed data at random could reduce  communication cost and strengthen the data privacy since adversaries are not able to regenerate the untransmitted region at the expense of performance. This partially uploaded smashed data is labelled as \textit{CutSmashed Data}. CutSmashed data conceal its information by random removal of patches. While SL requires to upload the entire smashed data of each client to the server, our proposed does not need to upload the whole smashed data from each client. 

Let there exist $n$ clients with a set of $\mathbb{C}=\{1,2,\cdots,n\}$ and a single server. 
$(\mathbf{x}_{i},\mathbf{y}_{i})$ denotes a batch of raw data-label tuples from $i$-th local dataset, $\mathbb{D}_i$. A neural network model is denoted with weights $\mathbf{w}_i$, which is splitted into two segments such that $\mathbf{w}_i=[\mathbf{w}_{c,i},\mathbf{w}_s]^\textrm{T}$ for $i \in \mathbb{C}$. We define $f$ as a representation for mapping from the input data to the output. The smashed data is expressed as $\mathbf{s}_{i}:=f_{\mathbf{w}_{c,i}}(\mathbf{x}_{i})$. Here, transformers divide the raw data into $M$ number of patches, and each patch is transformed into an embedding vector during FP. A smashed data is denoted as $\mathbf{s} = [e_1,...,e_M]\in \mathcal{R}^{M\times d_m}$, where $e_i$ is the $i$-th embedding vector and $d_m$ is the dimension of the patch embedding. 

Before uploading CutSmashed data, a piece of information which patch embeddings would be transmitted are shared between a server and clients. This prior information is called as \textit{a pseudo random sequence}, and is denoted as $\mathbb{B}$. Then, CutSmashed data is expressed as $\mathbf{s}'_i = \mathbb{B} \odot \mathbf{s}_i$, where $\mathbb{B}=[m_j]_{M\times 1}$, and  $m_j \in \{0,1\}$ where 0 indicates not to transmit, and 1 to transmit. $\odot$ denotes element-wise multiplication operation. Its communication cost can be ignored, since $\mathbb{B}$ is treated as a binary number which could be converted to one integer-type data having a negligible communication payload. \Cref{fig:examp2} shows examples of CutSmashed data whose black regions indicate the cut off regions. The pseudo random mask over the smashed could be created regardless of the depth of the cut layer since it is determined based on the number of patches of the smashed data.  

\begin{figure}[t]
\includegraphics[trim={0 0 0 1cm},clip, width=\columnwidth]{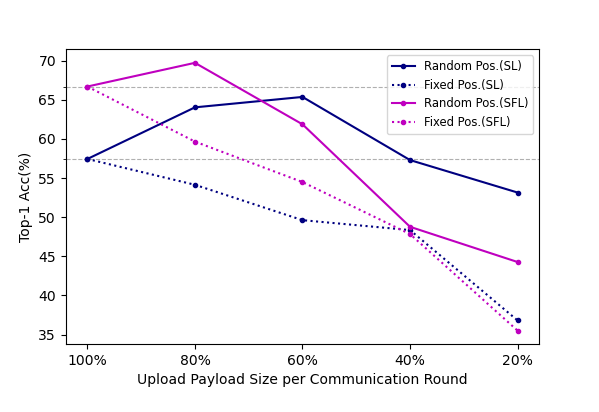}
\centering
\caption{Performance of SL with CutSmashed data w.r.t. the average size of CutSmashed data.}
\label{cutsmashed}
\end{figure}

\Cref{cutsmashed} shows the performance of SL and SFL training with CutSmashed data instead of smashed data in regard to the average size of CutSmashed data to identify the effect of the size of masking. The cases when the positions to be cut off are fixed for all rounds, and randomly selected at each iteration are compared. In the both cases, the performance decreases more with a larger size of the cutoff. The model cannot generalized well since the masked regions discard meaningful features of original data distribution. Interestingly, when the mask positions changed randomly, the performance is improved up to a certain extent of the cut off size. This implies that randomly generated CutSmashed data with a moderate size rather helps to prevent the model from overfitting to the local dataset acting as regularization like Cutout \cite{devries2017improved}. While the original Cutout putting one square region of mask to input, ours generate multiple masks to random locations. 

In terms of the leakage of data privacy, CutSmashed data is preferable than smashed data, since partial elements are concealed to the server. We elaborate on the data privacy leakage in the setting of a reconstruction attack where the attacker is willing to reconstruct an original data from uploaded smashed data. Hard to be restored by reconstruction attack implies raw data has strong privacy. In \Cref{Evaluation} evaluates the privacy leakage of CutSmashed data compared with the other techniques to be described in the next section.

\begin{figure}[t]
\includegraphics[width=0.7\columnwidth]{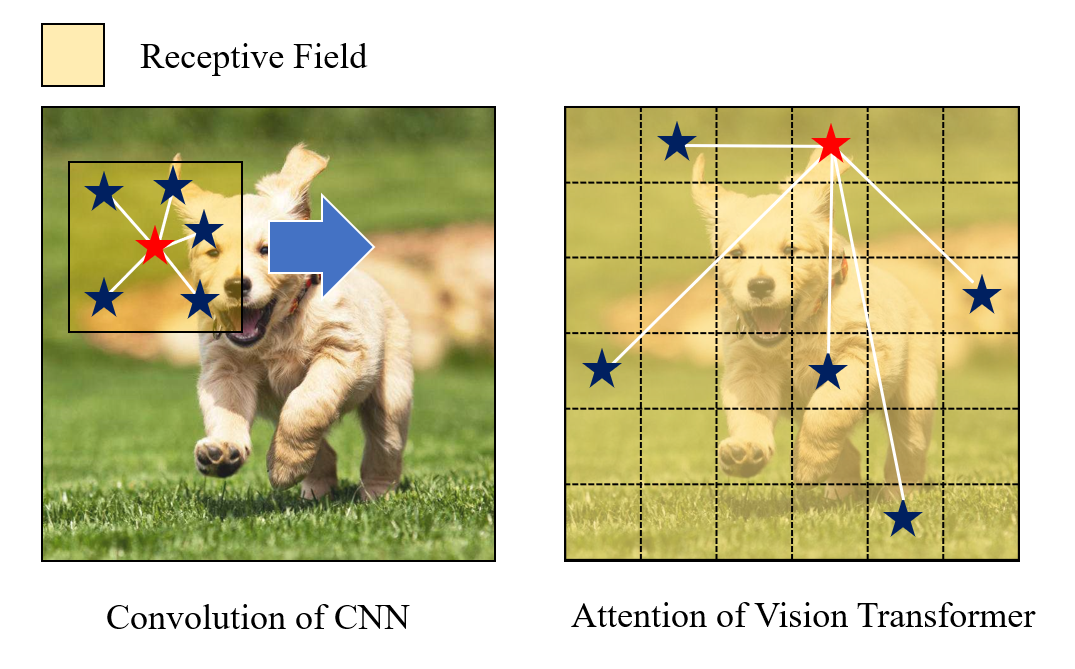}
\centering
\caption{Operation of CNN and ViT.}
\label{cnnvit}
\end{figure}


\subsection{Smashed Patch CutMix: Inter-client Mixup}
The vision transformer is difficult to be generalized well in situations where there is insufficient data held by clients due to its low inductive bias \cite{baxter2000model}.
Inductive bias is a set of assumptions added as prior information to solve unknown machine learning problems. The weak inductive bias causes reliance on data augmentation and model regularization to gather sufficient training data \cite{DBLP:journals/corr/abs-2106-10270}. Due to the data privacy, data augmentation between data held by different clients is limited in a distributed learning, making a severer problem for a transformer in a distributed learning. To overcome a limited inter-client data augmentation, and reduce the risk of the performance drop by uploading CutSmashed data, the blank parts of CutSmashed could be filled with patches of different clients' CutSmashed data. 

A self-attention mechanism evaluates which patches they should pay more attention to. Combining patches from different smashed data corresponds to a new attention between data from different clients. We demonstrate this assumption by putting CutSmashed data together from different clients during FP like putting the puzzle together. Since the process resembles CutMix \cite{yun2019cutmix} in that some region of data are cut and pasted into different data in the form of multiple patches with the same sizes, this operation is expressed as \textit{Smashed Patch CutMix}, and its mixed smashed data as \textit{CutMix data}. 
The central idea of \textit{Smashed Patch CutMix} is to create a new smashed data to supplement the performance loss due to partial uploading with filling blanks of CutSmashed data with each other. 
As shown in \Cref{fig:vit}, a virtual entity conducting the mix operation, \textit{a mixer}, is conceptualized to compare with SL and SFL. A mixer could be a 3rd party, or could be divided and belong to the server and the clients. Since overlapped patches lose its clear representation, non-overlapping mix between different CutSmashed data should be guaranteed, and paired pseudo random sequences are generated for Smashed Patch CutMix. 

We call $k$-way CutMix to represent that each $k$ number of smashed data are mixed to generate one CutMix data. Mixing groups, $\mathbb{G}=\{\mathbf{g}_1,..,\mathbf{g}_l\}$, are generated randomly from $\mathbb{C}$ at each iteration. Then, the number of patches allocated to each CutSmashed data (i.e. mixing ratio), $\{a_1, .., a_k\}$, is determined by sampling from a Dirichlet distribution with the dispersion factor $\alpha>0$ \cite{bishop2007discrete}.  Here, the number of trials of the distribution is set as the number $M$ of path tokens so that $\sum_{i=1}^{k}a_i = M$, and a probability vector is drawn from a Dirichlet distribution. Afterwards, patches at which positions of each smashed data are determined by the given mixing ratio. Here, the selected positions for patches of each client are not superposed with each other. Pseudo random sequence for $i$-th smashed data in a mixing group is denoted as $\mathbb{B}_{a_i}$, and $\mathbb{B}_{a_i}$ is determined uniformly and randomly.  $\sum_{j=1}^{M}m_{a_i,j}=a_i$ where $\mathbb{B}_{a_i}=[m_{a_i,j}]_{M\times 1}$ for $i \in [k]$. $\sum_{i=1}^{k}\mathbb{B}_{a_i}=\mathbf{1}_{M\times 1}$, where $\mathbf{1}$ is a vector of all ones.

Each client uploads CutSmashed data according to the received pseudo random sequence generated and shared by a mixer. The mixer conducts Smashed Patch CutMix, and CutMix data and its corresponding label by the 2-way Smashed Patch CutMix is expressed as:
\begin{align}
\tilde{\mathbf{s}}_{\{i,j\}}=\mathbf{s}'_i+\mathbf{s}'_j=\mathbb{B}_{a_i}\odot{\mathbf{s}_{i}}+\mathbb{B}_{a_j}\odot{\mathbf{s}_{j}}\\
\tilde{\mathbf{y}}_{\{i,j\}}=\frac{a_i}{M}\cdot{\mathbf{y}_{i}}+\frac{a_j}{M}\cdot{\mathbf{y}_{j}},
\label{eq:cutmix3}
\end{align}
where $\odot$ denotes the element-wise product.
\begin{algorithm}[t]
\caption{\small Operation of Mixer} 
\label{alg:main}
\begin{algorithmic}  \small
    \Function{Sequence Generation}{}
        \State {Sample \{$a_1,.., a_k$\} $\sim$ Dir($\alpha$) for $k$-way CutMix}
        \State {Generate  $\{\mathbb{B}_{a_1},..,\mathbb{B}_{a_k}\}$ uniformly at random}
        \State \Return {paired pseudo random sequences}
    \EndFunction
\State
      \For{$e \gets 1$ to $E$}
    \State {Generate a set of mixing groups, $\mathbb{G}$ from $\mathbb{C}$}
	\State \Call{Sequence Generation}{}
    \State {Send $\{\mathbb{B}_{a_1},.., \mathbb{B}_{a_k}\}$ to clients based on $\mathbb{G}$}
    \State {Receive $\{(\mathbf{s}'_1,\mathbf{y}_1),..,(\mathbf{s}'_n,\mathbf{y}_n)\}$ from $\mathbb{C}$}
	\For{$\mathbf{g} \in \mathbb{G}$} 
    	\State {$(\tilde{\mathbf{s}}_{\mathbf{g}}, \tilde{\mathbf{y}}_{\mathbf{g}}) \leftarrow ( \sum_{j\in\mathbf{g}}\mathbf{s}'_j,  \sum_{j\in\mathbf{g}}\frac{a_j}{M}\mathbf{y}_j ) $ }
        \State {Upload $\tilde{\mathbf{s}}_{\mathbf{g}}, \tilde{\mathbf{y}}_{\mathbf{g}}$ to the server}
	\EndFor
    \EndFor
    \end{algorithmic}
\end{algorithm}

The proposed is simple to implement with negligible extra communication cost for sharing pseudo random sequences. Intermixing CutSmashed data transmitted from each client's side enhances the server side's generalization capability acting as a data augmentation at feature space. Additionally, the impact of shuffling the order of the sequence on the performance is analyzed in detail in \Cref{Addon}. 

\begin{figure*}[t]
   \begin{subfigure}[ht]{0.33\textwidth}
    \includegraphics[trim={0 6cm 0cm 0},clip,width=\textwidth]{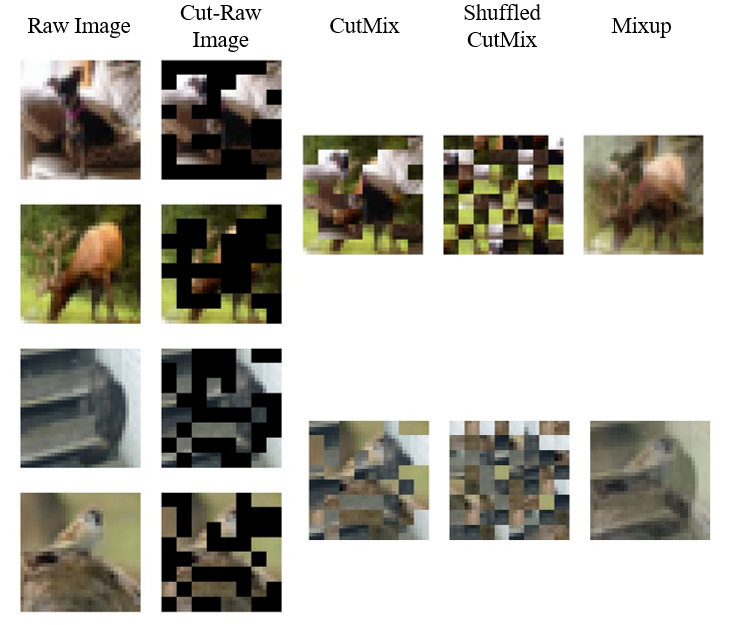}
    \centering
    \caption{Raw images.}
    \label{fig:examp1}
    \end{subfigure}
   \begin{subfigure}[ht]{0.33\textwidth}
    \includegraphics[trim={0 6cm 0cm 0cm},clip,width=\textwidth]{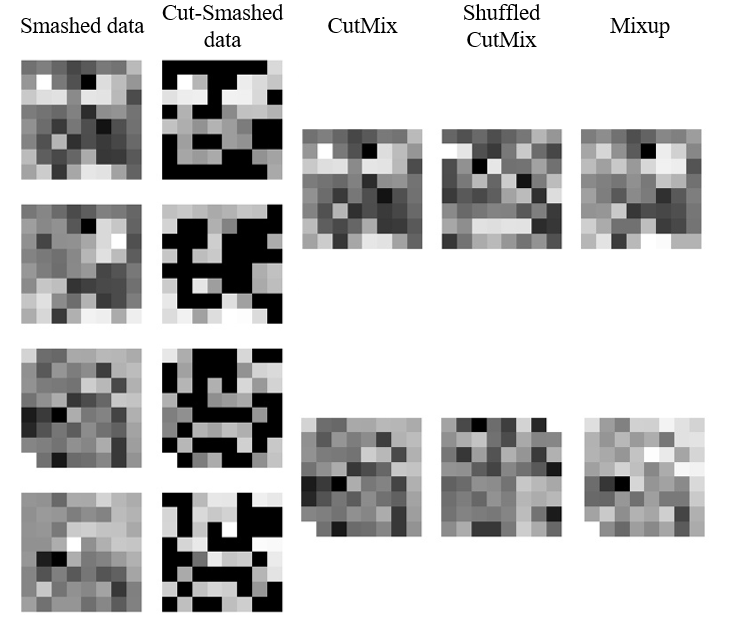}
    \centering
    \caption{Smashed data.}
    \label{fig:examp2}
    \end{subfigure}
   \begin{subfigure}[ht]{0.33\textwidth}
    \includegraphics[trim={0 6cm 0 0.1cm},clip,width=\textwidth]{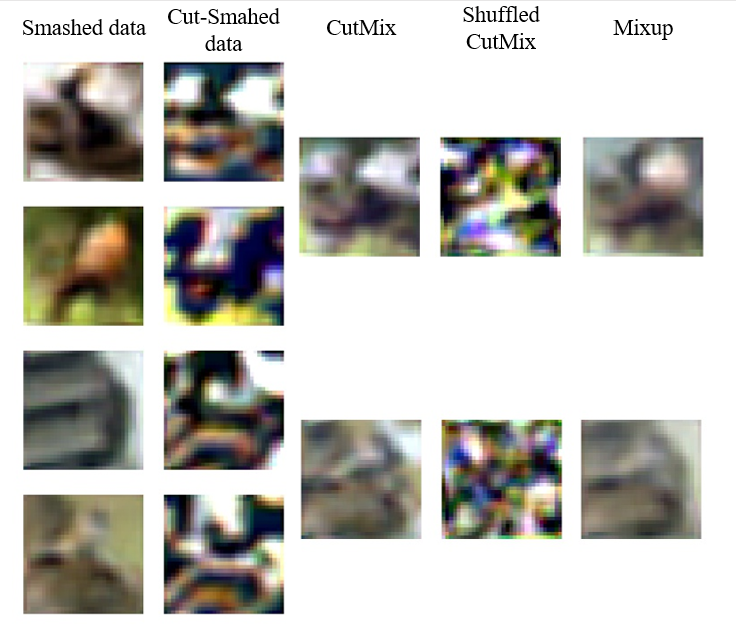}
    \centering
    \caption{Reconstruction from smashed data.}
    \label{fig:examp3}
    \end{subfigure}
    \caption{Examples of data with different operations at different levels.}
\end{figure*}

\subsection{Weight Update based on CutMix data}
At the server, the upper model segment $\mathbf{w}_{s}$ propagates $\tilde{\mathbf{s}}_{\{i,j\}}$ uploaded by $i$-th and $j$-th client, and generates softmax output $f_{\mathbf{w}_{s}}(\mathbf{s}_{\{i,j\}})$. 
Then, the loss $\tilde{L}_{\{i,j\}}$ generated by the server model with CutMix data can be expressed as:
\begin{align}
    \tilde{L}_{\{i,j\}}= \frac{1}{b}\sum \text{CE}\!\left(f_{\mathbf{w}_{s}} ( \tilde{\mathbf{s}}_{\{i,j\}} ),\tilde{\mathbf{y}}_{\{i,j\}}\right), \label{Eq:smashFP2}
\end{align} 
where $b$ is a batch size. 
The weights of the server and the clients are updated by BP as follows:
\begin{align}
\begin{bmatrix}
\mathbf{w}_s \\  \mathbf{w}_{c, i\in \mathbb{C}}
\end{bmatrix}
\leftarrow 
\begin{bmatrix}
\mathbf{w}_s \\  \mathbf{w}_{c, i\in \mathbb{C}}
\end{bmatrix}
-\eta
\begin{bmatrix}
\sum_{\{i,j\}\in{\mathbb{G}}}{(\nabla_{\mathbf{w}_s}\tilde{L}_{\{i,j\}})}\\
\nabla_{\mathbf{w}_{c,i}}\tilde{L}_{\{i,j\}} 
\end{bmatrix}
\end{align}
where $\mathbb{G}$ is a set of groups of clients whose smashed data are mixed, $\eta$ is a learning rate, and $\nabla_{\mathbf{w}_s}\tilde{L}_{\{i,j\}}$ and  $\nabla_{\mathbf{w}_{c,i}}\tilde{L}_{\{i,j\}}$ are the derivatives of the error with respect to $\mathbf{w}_s$ and $\mathbf{w}_{c,i}$, respectively. 
The server sends the gradients of $\tilde{L}_{i,j}$ with respect to the uploaded smashed data $\mathbf{s}'_i$ and $\mathbf{s}'_j$, $\nabla_{\mathbf{s}_i} \tilde{L}_{i,j}$ and $\nabla_{\mathbf{s}_j} \tilde{L}_{i,j}$, to the corresponding clients, and the clients calculates $\nabla_{\mathbf{w}_{c,i}}\tilde{L}_{\{i,j\}} $, which is given as:
\begin{align}
\nabla_{\mathbf{w}_{c,i}}\tilde{L}_{i,j} 
=\frac{\partial\tilde{L}_{\{i,j\}}}{\partial \mathbf{s}'_i}\frac{\partial \mathbf{s}'_i}{\partial \mathbf{w}_{c,i}}.
\label{eq:gradientx}
\end{align}

The method above aims to resolve a server-clients update imbalance problem, hindering the scalability of parallel split learning \cite{oh2022locfedmix}. The more clients, it incurs the more imbalanced updates between the server and the clients. There are $n$ times of update on the upper model segment, whereas each client updates its model once. Meanwhile, with our proposed method, the server executes its server model's update only once on a unit mixed smashed data, as shown in \Cref{fig:vit}. Therefore, when all clients update their parameters with CutMix data, $n$ times updates are in the server model reduce to $\frac{n}{k}$ times updates for $k$-way CutMix. The CutMixSL ($k$ times) in \Cref{scalability} is the case when gradients from one CutMix data flow to only one client and the server updates $n$ times in total like SL. Compared with CutMixSL in \Cref{scalability} where the server updates $\frac{n}{k}$ times, it shows that the reduced update of the server by CutMix data has a positive impact on a performance gain.  

\section{Evaluation} \label{Evaluation}

We consider the CIFAR-10 (60k samples of 32x32 pixel color images with 10 classes) classification task with up to 10 clients, where each client stores randomly selected 5k samples and is associated with a common server. The standard ViT architecture is visualized in Figure 5a. In our setting, we follow the ViT-Tiny architecture \cite{DBLP:journals/corr/abs-2012-12877} including 6 Transformer blocks, each of which consists of multiple layers for operating the attention mechanism. The ViT-Tiny is split into two segments, and the cut-layer is the embedding layer, i.e., the embedding layer is stored at each client while the rest is offloaded to the server. Given the CIFAR-10 dataset, each sample is divided into 4x4 patches that are flattened into a sequence of 16 tokens. The embedding layer concatenates these patch tokens and the class token (i.e., the sample's class), followed by adding positional embedding to identify the patch positions within the sample. To train this paralle split ViT, we use the adamW optimizer, and the learning is 0.001 with the cosine annealing. The batch size is 128, and the total number of epochs is 600 with 5 warmup epochs.

\begin{figure}[t]
\centering
\includegraphics[trim={0 0 0 0},clip, width=0.8\columnwidth]{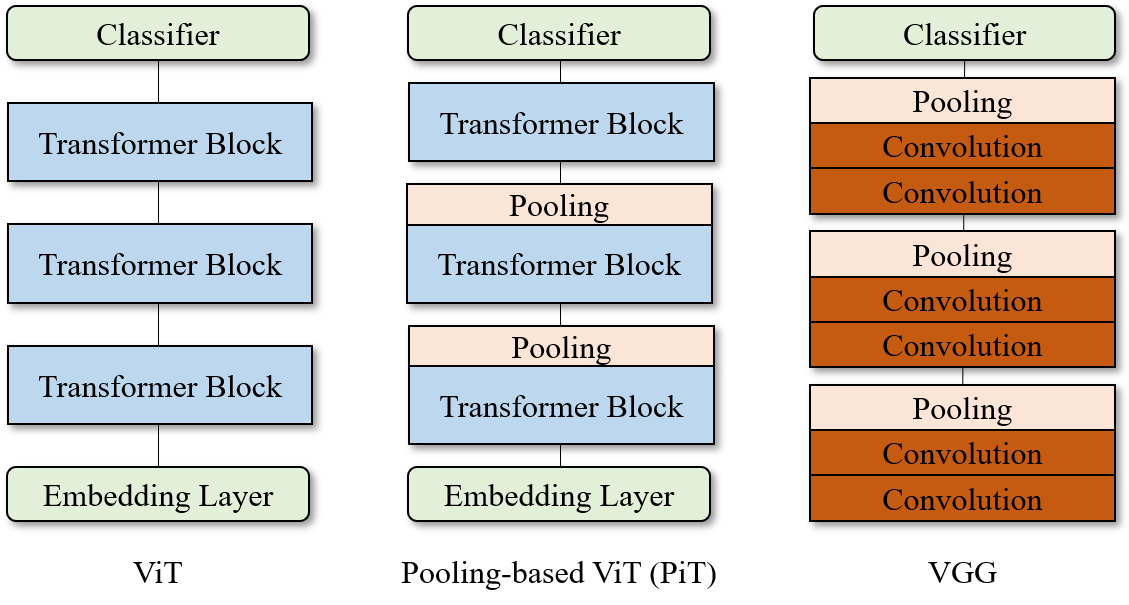}
\caption{Simplified model architectures of ViT, PiT, and VGG.}
\label{modelarhitecture}
\end{figure}

\paragraph{Impact of Mixing Methods}
To analyze the effectiveness of Smashed Patch CutMix on the transformer, different types of mix operations on both transformer-based models and a CNN-based model are evaluated. ViT-Tiny for a pure transformer, PiT-Tiny \cite{DBLP:journals/corr/abs-2103-16302} for a pooling-based transformer, and VGG-16 \cite{simonyan2014very} for CNN are used whose simplified model architectures are described in \Cref{modelarhitecture}.
Many ViT models with hierarchical representations are proposed such as Swin Transfromer \cite{liu2021swin}, and T2T-ViT \cite{yuan2021tokens} including PiT. VGG is a representative model of CNN composed of convolutional layers, pooling layers and a classifier, and the cut-layer is after two convolutional and one pooling layers.

\begin{table}[t]
\centering
    \resizebox{0.65\columnwidth}{!}{
\begin{tabular}{l|rrrr}
\toprule
\multirow{2}{*}{Method} &  \multicolumn{3}{c}{Models}\\ \cmidrule(l){2-4}
 & ViT-Tiny & PiT-Tiny & VGG-16  \\ \midrule
Standalone & 49.14 & 47.77 & 54.97\\
Parallel SL  & 57.05 & 52.28 & 62.62 \\\midrule 
Cutout & 65.03 & 60.87 & 67.06 \\ 
Mixup & 71.02 &65.92  & \textbf{74.43} \\ 
Patch CutMix  & \textbf{75.55} &  \textbf{73.19} &72.23\\
Shuffled CutMix  & 72.78 & 57.59 & 33.50 \\ 
\bottomrule
\end{tabular}}
\caption{Performance w.r.t. mixing methods.}
\label{table:models}
\end{table}

\Cref{table:models} shows the performances of Cutout (CutSmashed data), Mixup (Smashed data + Mixup), Smashed Patch CutMix (CutSmashed data + CutMix), and Shuffled CutMix (CutSmashed data + CutMix + Shuffling). Smashed Patch CutMix has 18.5\% and 20.9\% performance gain on ViT and PiT compared to parallel SL. It is higher than Mixup for ViT and PiT (4\% and 8\%, respectively), and 2\% lower than Mixup for VGG-16. Masking such as CutMix is better at preserving local features than interpolation such as mixup without distorting data distribution \cite{DBLP:journals/corr/abs-2002-12047}. Transformers process data as divided separate patches, and even though randomly chosen patches to be cut and pasted destroy a global structure of image, transformers do not lose its inference ability because of parallel processing for an entire range of sequence by attention mechanism. In this sense, masking smashed data is expected to be more suitable than interpolation to transformers. On the other hand, CNN uses sliding convolutional filters and one of assumptions CNN poses is a locality of pixel dependencies \cite{krizhevsky2012imagenet}. Neighboring pixels which tend to be correlated, get uncorrelated by masking due to junctions of different clients' patches. Notwithstanding, the result shows that the impact of masking is not big enough to degrade the performance of CNN. We guess that the unchanged position of patches from the original location keeping its architecture of the data results in a positive impact to the performance gain. 

\begin{figure}[t]
\centering
\begin{subfigure}{0.45\columnwidth}
\includegraphics[trim={0.35cm 0 0.7cm 0.3cm},clip, width=\columnwidth]{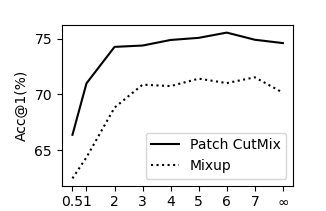}
\centering
\caption{Mixing ratio.}
\label{fig:alphavalue}
\end{subfigure}
\hfill
\begin{subtable}{0.5\columnwidth}
        \centering
        \resizebox{.8\columnwidth}{!}{
        \begin{tabular}{c|cc}
        \toprule
        \multirow{2}{*}{Type} &  \multicolumn{2}{c}{Mixing Ratio}\\ \cmidrule(l){2-3}
          & Uniform & Dir($\alpha=6$)\\ \cmidrule(l){1-3}
        2-way & \textbf{70.93}   & \textbf{75.55}\\ 
        3-way & 68.59  & 70.86\\ 
        4-way & 66.96   & 66.67\\ 
        5-way & 67.01& 67.04 \\ 
        \bottomrule
        \end{tabular}}
\caption{Mixing group size.}
\label{table:way}
\end{subtable}    
\caption{Top-1 Accuracy w.r.t. (a) mixing ratio $\alpha$ and (2) group size.}
\label{ablation}
\end{figure}

\paragraph{Impact of Mixing Ratio \& Group Size} In \Cref{fig:alphavalue}, the performance of CutMixSL with respect to $\alpha$ of Dirichlet-multinomial distribution is evaluated to identify a sensitivity to the mixing ratio. The higher $\alpha$, the higher the probability of even mixing ratio. The result shows that the higher probability of evenly mix gains more accuracy than the lower one. The top-1 accuracy when $\alpha \rightarrow \infty$, which is the number of sampled patches is fixed to even, is lower than the case for the lower $\alpha$. It implies that having an uneven mixture with a moderate probability is preferable for a higher performance. \Cref{table:way} evaluated a performance of the proposed in regard to the size of mixing group with a uniform distributions and the dirichlet distribution with $\alpha=6$. It is expected that the performance would be improved with the larger mixing group size from the standpoint of update imbalance problem, since the server updates less. Nevertheless, 2-way CutMix has a best performance gain implying that overly mixing with different classes causes the server cannot learn distinct features of each class.

\paragraph{Scalability}
In \Cref{scalability}, while all evaluated methods are scalable up to at least 10 clients, each has a different performance gain at scale. The brackets in \Cref{scalability} represents the distribution that a mixing ratio is sampled from. Parallel SL has a weak scalable performance gain {\small (4\%)} due to the update imbalance problem, whereas CutMixSL has a higher scalable performance gain than parallel SL {\small(12.9\%)} and even SFL {\small(12.7\%)}, and an top-1 accuracy of CutMixSL is also higher than SFL. CutMixSFL, which is CutMixSL with FedAvg on the lower models, achieves 13.4\% performance gain at scale and about 80\% top-1 accuracy with ten clients. 

\begin{table}[t]
\centering
\resizebox{\columnwidth}{!}{
\begin{tabular}{l|rrrrr}
\toprule
\multirow{2}{*}{Method} & \multicolumn{5}{c}{$\#$ of clients}\\ \cmidrule(l){2-6}
& 2 & 4& 6 & 8&  10\\ \cmidrule(l){1-6}
Parallel SL  &  \gradiento{52.91} & \gradiento{55.55} & \gradiento{56.81} & \gradiento{56.55 }& \gradiento{57.05}\\
SplitFed & \gradienttw{53.96} & \gradienttw{58.39} & \gradienttw{63.07} &\gradienttw{65.59} &\gradienttw{66.70}\\
CutMixSL {\footnotesize($k$ times, uniform)} & \gradientt{58.06} & \gradientt{65.11} & \gradientt{65.76} & \gradientt{68.97} & \gradientt{67.66}\\ 
CutMixSL {\footnotesize(uniform)} & \gradientf{61.66} & \gradientf{66.80} & \gradientf{67.69} & \gradientf{69.86} &\gradientf{70.93}\\ 
CutMixSL {\footnotesize($\alpha$=6)} & \gradientf{62.62} & \gradientf{69.47} & \gradientf{71.80} & \gradientf{73.67} & \gradientf{75.55}\\
CutMixSFL {\footnotesize(uniform)} & \gradientfi{65.82} & \gradientfi{72.31} & \gradientfi{74.31} & \gradientfi{78.88} &\gradientfi{80.14}\\ 
CutMixSFL {\footnotesize($\alpha$=6)} & \gradientfi{67.58} & \gradientfi{73.52} & \gradientfi{76.85} & \gradientfi{79.48} &\gradientfi{80.97}\\
\bottomrule
\end{tabular}}
\caption{Scalability. Top-1 accuracy w.r.t. the \# of clients.}
\label{scalability}
\end{table}

\begin{figure}[t]
\includegraphics[trim={0 0 0cm 1.3cm},clip, width=0.9\columnwidth]{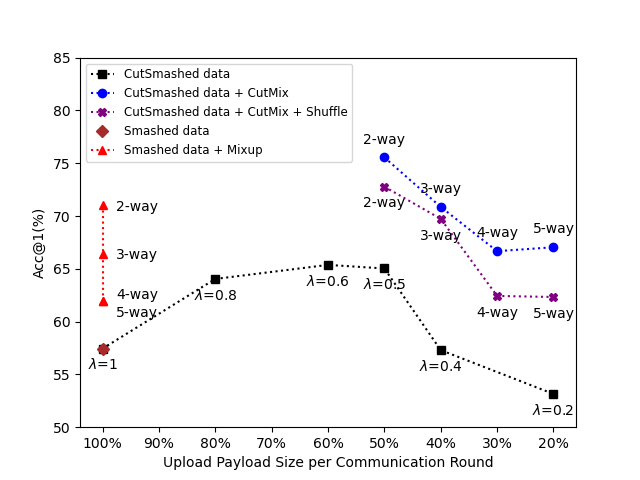}
\centering
\caption{Upload payload size per communication round of CutMixSL and its derivatives.}
\label{fig:commeffif}
\end{figure}

\paragraph{Communication Efficiency.}
Compared to Parallel SL requiring entire smashed data, one of the advantages of the proposed technique is a large reduction of upload payload size with an enhanced accuracy. \Cref{fig:commeffif} shows an upload payload size per communication round and the performance of CutMixSL and its derivatives. The black line shows an accuracy of SL with CutSmashed data regarding $\lambda$, the ratio of the size of CutSmashed data to the one of the original smashed data described in \Cref{cutsmashed}. CutMixSL shows the highest performance and upload payload reduction up to 20 - 50\%. The more mixing group size increases, the higher communication cost reduction can be achieved with a smaller performance gain. In our settings, one unmatched client of $3$-way sends smashed data, and two unmatched clients in 4-way CutMix conduct 2-way CutMix.       

\paragraph{Privacy Leakage}

Data privacy leakage increases with the mutual information between the CutSmashed data and its raw data. Due to the unknown data distributions, the mutual information is often approximated by the error when reconstructing the input data \cite{vepakomma2020nopeek}. In this respect, following \cite{wang2021revisiting}, we train an autoencoder whose input is the CutSmashed data and the output is the original raw data. The trained autoencoder's loss, given as the mean squared error (MSE), can thereby treated as the amount of privacy leakage to the server.


\Cref{talbe:recon} shows the privacy leakage of 2-way CutMixSL and its variants on a test set when the autoencoder trains 10\% of a train set, and when it trains 100\% of a train set. As intuitively expected, training with CutSmashed data reduces privacy leakage by 15x compared to the one with smashed data. For the mix operations, Smashed Patch CutMix has higher reconstruction loss than Mixup, and CutMixSL reduces data privacy by around 8 times compared to the baseline. This gap gets highly amplified when the order of its sequences is shuffled (Shuffled CutMix), implying that masking data and shuffling are two principal factors for privacy enhancement. \Cref{fig:examp3} shows some examples of reconstructed images by the reconstruction attack for a qualitative comparison. 

\begin{table}[t]
\centering
    \resizebox{7cm}{!}{
\begin{tabular}{l|rr}
\toprule
Type &   Train Dataset(10\%) & Train Dataset(100\%) \\ \cmidrule(l){1-3}
Smashed data & 0.0091 & 0.0056\\ 
CutSmashed data & \textbf{0.0920} & \textbf{0.0829}\\ \cmidrule(l){1-3}
Mixup & 0.0402 & 0.0351\\ 
Patch CutMix & 0.0458 & 0.0434\\  
Shuffled CutMix & \textbf{0.1233} & \textbf{0.1250}\\ 
\bottomrule
\end{tabular}}
\caption{Privacy leakage measured by the reconstruction loss (MSE).}
\label{talbe:recon}
\end{table}

\section{Conclusion}
In this work, CutSmashed data is introduced to resolve data privacy leakage and to improve communication efficiency of split learning motivated by a process of a vision transformer handling data as a sequence of patches. Furthermore, CutMixSL is proposed with Smashed patch CutMix, a smashed data augmentation, to deploy a transformer-based model for split learning. We analyzed the design elements' impact on the proposed operation, and confirmed that the it has advantages on a improved performance, communication efficiency, and data privacy compared to Parallel SL and SFL. \medskip
\clearpage

\section*{Acknowledgments}
This work was partly supported by Institute of Information \& communications Technology Planning \& Evaluation (IITP) grant funded by the Korea government (MSIT) (No.2021-0-00347, 6G Post-MAC (POsitioning- \& Spectrum-aware intelligenT MAC for Computing \& Communication Convergence)), and Institute of Information \& communications Technology Planning \& Evaluation (IITP) grant funded by the Korea government(MSIT) (1711134613, 2021-0-00270, Development of 5G MEC framework to improve food factory productivity, automate and optimize flexible packaging).

\bibliographystyle{named}
\bibliography{egbib.bib}

\appendix
\clearpage
\section{Pseudo Algorithm of CutMixSL}
\begin{algorithm}[h]
\caption{CutMixSL} 
\label{alg:appen}
\begin{algorithmic}
    \Function{Sequence Generation}{}
        \State {Sample \{$a_1,.., a_k$\} $\sim$ Dir($\alpha$) for $k$-way CutMix}
        \State {Generate  $\{\mathbb{B}_{a_1},..,\mathbb{B}_{a_k}\}$ uniformly at random}
        \State \Return {paired pseudo random sequences}
    \EndFunction
\State
      \For{$e \gets 1$ to $E$}
    \State {\textbf{/*Runs on mixer*/}}
    \State {Generate a set of mixing groups, $\mathbb{G}$ from $\mathbb{C}$}
	\State \Call{Sequence Generation}{}
    \State {Send $\{\mathbb{B}_{a_1},.., \mathbb{B}_{a_k}\}$ to clients based on $\mathbb{G}$}
    \State {Receive $\{(\mathbf{s}'_1,\mathbf{y}_1),..,(\mathbf{s}'_n,\mathbf{y}_n)\}$ from $\mathbb{C}$}
	\For{$\mathbf{g} \in \mathbb{G}$} 
    	\State {$(\tilde{\mathbf{s}}_{\mathbf{g}}, \tilde{\mathbf{y}}_{\mathbf{g}}) \leftarrow ( \sum_{j\in\mathbf{g}}\mathbf{s}'_j,  \sum_{j\in\mathbf{g}}\frac{a_j}{M}\mathbf{y}_j ) $ }
        \State {Upload $\tilde{\mathbf{s}}_{\mathbf{g}}, \tilde{\mathbf{y}}_{\mathbf{g}}$ to the server}
	\EndFor
\State{}
    \State {\textbf{/*Runs on clients*/}}
    \For{each client $i \in \mathbb{C}$ in parallel}
    \State {Receive $\mathbb{B}_i$ from mixer}
    \State {$\mathbf{s}_i \leftarrow f_{\mathbf{w}_{c,i}}(\mathbf{x}_i); \mathbf{s}'_i \leftarrow \mathbb{B}_i \odot \mathbf{s}_i$}
    \State {Upload $(\mathbf{s}'_i, \mathbf{y}_i)$ to mixer}
    \State {Receive $\nabla_{\mathbf{s}_i}{\tilde{L}_{\mathbf{g}}}$; Calculate gradient $\nabla_{\mathbf{w}_{c,i}}\tilde{L}_{\mathbf{g}} $}
    \State {Weight Update $\mathbf{w}_{c,i} \leftarrow \mathbf{w}_{c,i} -\eta \nabla_{\mathbf{w}_{c,i}}\tilde{L}_{\mathbf{g}} $}
    \EndFor
    \State {}
    \State {\textbf{/*Runs on server*/}}
    \For{$\mathbf{g} \in \mathbb{G}$} 
    \State {Receive $(\tilde{\mathbf{s}}_{\mathbf{g}}, \tilde{\mathbf{y}}_{\mathbf{g}})$ from mixer}
    \State {$\tilde{L}_{\mathbf{g}} \leftarrow \frac{1}{b}\sum_{\{\tilde{\mathbf{s}}_{\mathbf{g}},\tilde{\mathbf{y}}_{\mathbf{g}}\}} \text{CE}\!\left(f_{\mathbf{w}_{s}} ( \tilde{\mathbf{s}}_{\mathbf{g}} ),\tilde{\mathbf{y}}_{\mathbf{g}}\right)$}
    \State {Calculate gradient $\nabla_{\mathbf{w}_{s}}\tilde{L}_{\mathbf{g}} $ }
    \State {Weight Update $\mathbf{w}_{s} \leftarrow \mathbf{w}_{s}-\eta\nabla_{\mathbf{w}_{s}}\tilde{L}_{\mathbf{g}} $}
    \State {Download $\nabla_{\mathbf{s}_i}{\tilde{L}_{\mathbf{g}}}$ to client $i \in \mathbf{g}$ }
    \EndFor
    \EndFor
    \end{algorithmic}
\end{algorithm}



\section{Mixing Methods}\label{Mixing Method}
Data augmentation can generate new samples through mixing different samples, and there are generally two types of inter-sample data augmentation: masking and interpolation. The typical techniques of masking are CutMix \cite{yun2019cutmix} and Cutout \cite{devries2017improved}, and the one of interpolation is Mixup \cite{zhang2017mixup}.

Mixup is an entire interpolation of two given raw data, $x_i$ and $x_j$ with a certain ratio $\lambda$, and is expressed as follows:

\begin{equation}
\begin{gathered}
\mathbf{x}_{mixup}=\lambda{\mathbf{x}_{i}}+(1-\lambda){\mathbf{x}_{j}}.
\end{gathered}
\label{eq:mixup}
\end{equation}

In \cite{verma2019manifold}, Manifold Mixup has been proposed, which is Mixup at the feature space of the deep neural network (DNN), and is expressed as follows: 

\begin{equation}
\begin{gathered}
\mathbf{s}_{mixup}=\lambda{\mathbf{s}_{i}}+(1-\lambda)\mathbf{s}_{j}.
\end{gathered}
\label{eq:mani_mixup}
\end{equation}

The corresponding label for the generated data by Mixup and Manifold Mixup is expressed as follows:  
\begin{equation}
\begin{gathered}
\mathbf{y}_{mixup}=\lambda{\mathbf{y}_i}+(1-\lambda){\mathbf{y}_j}.
\end{gathered}
\end{equation}

As a masking technique, Cutout \cite{devries2017improved} masks out square regions of input, and is expressed as follows:

\begin{equation}
\begin{gathered}
\mathbf{x}_{cutout}=\mathbf{M}\odot{\mathbf{x}},
\end{gathered}
\end{equation}
where $\mathbf{M} \in \{0,1\}^{W\times H}$ is a binary mask indicating which pixels are to be dropped out. $\mathbf{M}$ fills $0$ inside the bounding box coordinates, $\mathbf{B}=(r_x, r_y,r_w,r_h)$, indicating the cropping region of the image. 

As one step further, CutMix drops a unit square region of random size, and fills in the blanks with a different raw image, and is expressed as following:
 
\begin{equation}
\begin{gathered}
\mathbf{x}_{cutmix}=\mathbf{M}\odot{\mathbf{x}_i}+(1-\mathbf{M}){\mathbf{x}_j}\\
\mathbf{y}_{cutmix}=\lambda{\mathbf{y}_i}+(1-\lambda){\mathbf{y}_j},
\end{gathered}
\end{equation}
where $\lambda = \frac{r_w r_h}{WH}$, is a mixing ratio. 

Smashed Patch CutMix, which is proposed in this paper, is operated in the feature space and is expressed as follows: 
\begin{equation}
\begin{gathered}
\mathbf{s}_{cutmix}=\mathbf{M}\odot{\mathbf{s}_i}+(1-\mathbf{M}){\mathbf{s}_j}\\
\mathbf{y}_{cutmix}=\lambda{\mathbf{y}_i}+(1-\lambda){\mathbf{y}_j}
\end{gathered}
\end{equation}

The proposed masks a random number of patches with other client's patches, which is similar to the CutMix except for the number of patches to be replaced; ours use multiple fixed-size patches.

Shuffled CutMix conducts shuffling after mixing smashed data by Smashed patch CutMix, and is expressed as follows: 

\begin{equation}
\begin{gathered}
\mathbf{s}_{shuffle}=\mathbf{S}(\mathbf{s}_{cutmix}),
\end{gathered}
\end{equation}
where $\mathbf{S}$ is a shuffle operation on $\textbf{s}_{cutmix}= [e_1,...,e_M]\in \mathcal{R}^{M\times d_m}$. $e_i$ is the $i$-th embedding vector, $d_m$ is the size of a vector of a patch, and $\mathbf{S}$ shuffles the sequence of the embedding vectors of $\mathbf{s}_{cutmix}$. For CNN, activations can be reshaped to a 2D dimension like activations in a vision transformer by dividing it with a given patch size, and aligning them in parallel.     

\section{Add-On Experiments}\label{Addon}

\paragraph{Shuffling} 
Transformers process data as a long-range sequence in parallel by attention mechanism, and are not heavily affected by sequence order. This high permutation invariance also applies to ViTs \cite{naseer2021intriguing}, and could bring benefit on reducing privacy leakage by reconstruction attacks in SL since shuffling patches can destroy images' overall structure as shown in \Cref{fig:examp1}. While all patches retain their own positions during mixing by Smashed Patch CutMix, it can be extended to a shuffled version, \textit{Shuffled CutMix}, in short, for a further privacy enhancement. 

The last row of \Cref{table:models} shows that the influence of shuffling an order of patches of CutMix data. ViT keeps its performance in spite of shuffling with a slight performance loss by its capability of high permutation invariance. However, PiT has a pooling operation that blurs shuffled patches and lose their own distinct features, and this tendency of a degenerated performance is more serious for CNN due to a consequence of convolutional filters in addition to pooling. Nevertheless, a considerable profit by shuffling is an improvement of reconstruction mitigation as shown in \Cref{talbe:recon}. It reduces the privacy leakage by around 22 $\times$ compared to the the baseline (SL with smashed data) with the ViT model.

\begin{figure}[t]
\centering
\begin{subfigure}[h]{0.238\textwidth}
\centering
\includegraphics[trim={0 0 12cm 0},clip,width=\textwidth]{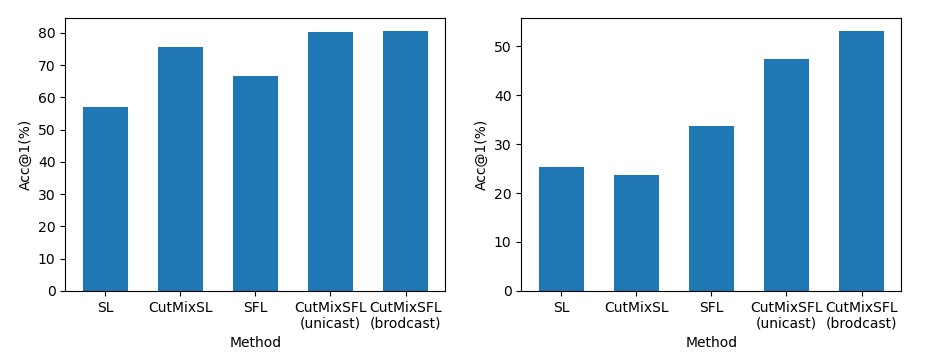}
\caption{IID.}
\label{fig:iid}
\end{subfigure}
\hfill
\begin{subfigure}[h]{0.238\textwidth}
\centering
\includegraphics[trim={12cm 0 0 0},clip,width=\textwidth]{images/nonIID.png}
\caption{non-IID($\mu=0.1$).}
\label{fig:noniid}
\end{subfigure}
\caption{Performance w.r.t. IID and non-IID data distribtuion.}
\label{datadistribution}
\end{figure}

\paragraph{FedAvg}
FedAvg on clients can be utilized to enhance the performance gain and to be tolerant to data distribution shifts. When data are non-IID to clients, each client contains one or two dominant classes' data, training its one-sided classes predominantly. A mix operation like Smashed Patch CutMix can alleviate unbalanced training in cooperation with FedAvg. In \Cref{fig:noniid}, dirichlet distribution with concentration parameter, $\mu$, 0.1 is used to formulate the non-IID case. The result indicates that the degenerated performance of SL and SFL by non-IID condition can be less worsened by Smashed Patch CutMix and FedAvg. 

During BP through CutMix data, the flow of gradients to each client can be calculated from the perspective of the server and and the perspective of clients. \cite{pal2021server} uses local gradient averaging for broadcasting the gradients to clients. Likewise, the gradients from CutMix data can flow as unicast and broadcast.

\begin{align}
\nabla_{\mathbf{w}_{c,i}}\tilde{L}_{i,j} 
\approx
\begin{cases}
\frac{\partial\tilde{L}_{\{i,j\}}}{\partial \mathbf{s}'_i}\frac{\partial \mathbf{s}'_i}{\partial \mathbf{w}_{c,i}};& \textrm{for unicast}\\
\frac{\partial\tilde{L}_{\{i,j\}}}{\partial \mathbf{s}_{\{i,j\}}}\frac{\partial \tilde{\mathbf{s}}_{\{i,j\}}}{\partial \mathbf{w}_{c,i}};& \textrm{for broadcast}
\end{cases}.
\label{eq:gradient}
\end{align}

The unicast case is that the gradients is calculated in the perspective of clients, and the server knows paired pseudo random sequences indicating which patches each client uploaded. The server sends individual gradients, $\nabla_{\tilde{s}_{i}}\tilde{L}_{i,j}$ and $\nabla_{\tilde{s}_{j}}\tilde{L}_{i,j}$, to $i$-th and $j$-th clients respectively according to the corresponding portion of CutMix data by unicast. For the broadcast case, the server is assumed not to know the pseudo random sequences, thereby the gradient with respect to the whole region of CutMix data is sent during BP. It is named as \textit{CutMix Gradient}, and it broadcasts combined gradients, $\nabla_{\tilde{s}_{\{i,j\}}}\tilde{L}_{i,j}$, to the clients whose CutMix data are generated from. CutMix Gradient could be used when the server does not know the pseudo random sequences. 

Although the difference of the performances is negligible in the IID condition, the performance of CutMixSFL with CutMix gradients is 6\% higher than the one with the gradients with respect to CutSmashed data in the non-IID condition. It could be interpreted that CutMix Gradients by the mixed smashed data is more effective in that each model trains data of scarce classes indirectly through CutMix gradient with the help of FedAvg.

\paragraph{Noise Injection}
   
\begin{figure}[t]
\includegraphics[trim={0 0 0 0cm},clip, width=0.7\columnwidth]{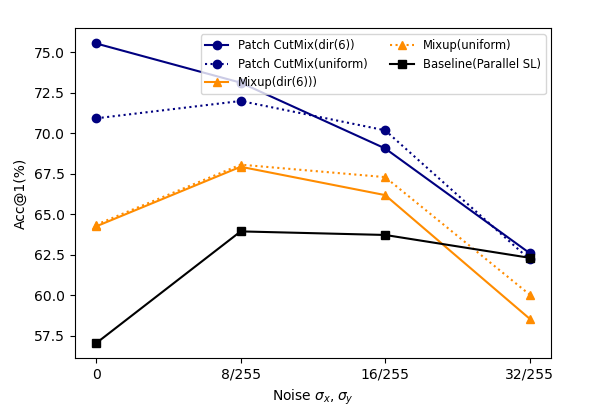}
\centering
\caption{Utility according to the scale of gaussian noise added to the smashed data and its label.}
\label{fig:DPutil}
\end{figure}

Many privacy-preserving data mining techniques involve noise injection, such as differential privacy, to randomly distort and mask data reducing the distance correlation between an output of a mechanism and a raw input data. In \Cref{fig:DPutil}, the utility of mixing operations are evaluated when additive white gaussian noises are integrated to the smashed data and its corresponding label for data privacy. It shows a comparison of utility of Smashed Patch CutMix, Smashed Mixup, and the baseline (Parallel SL) according to the scale, $\sigma_x$ and $\sigma_y$ of the noise distribution on the raw data and its label, respectively. The higher $\sigma$ preserves stronger privacy at the larger cost of utility. The result indicates the proposed has a higher utility compared to the others, even though the gap is decreased with a higher $\sigma$. One interesting founding is that two cases except for the proposed show utility gains when a small amount of noise is injected in \Cref{fig:DPutil}. It could be explained that two methods get a positive effect through regularization making the model less certain of its predictions, while the proposed does not so since it already gets an enough regularization and data augmentation effect. 

\section{Additional Visualization of images}\label{appen_images}
\begin{figure*}[ht]
   \begin{subfigure}[ht]{0.33\textwidth}
    \includegraphics[trim={0 0 27cm 0},clip,width=\textwidth]{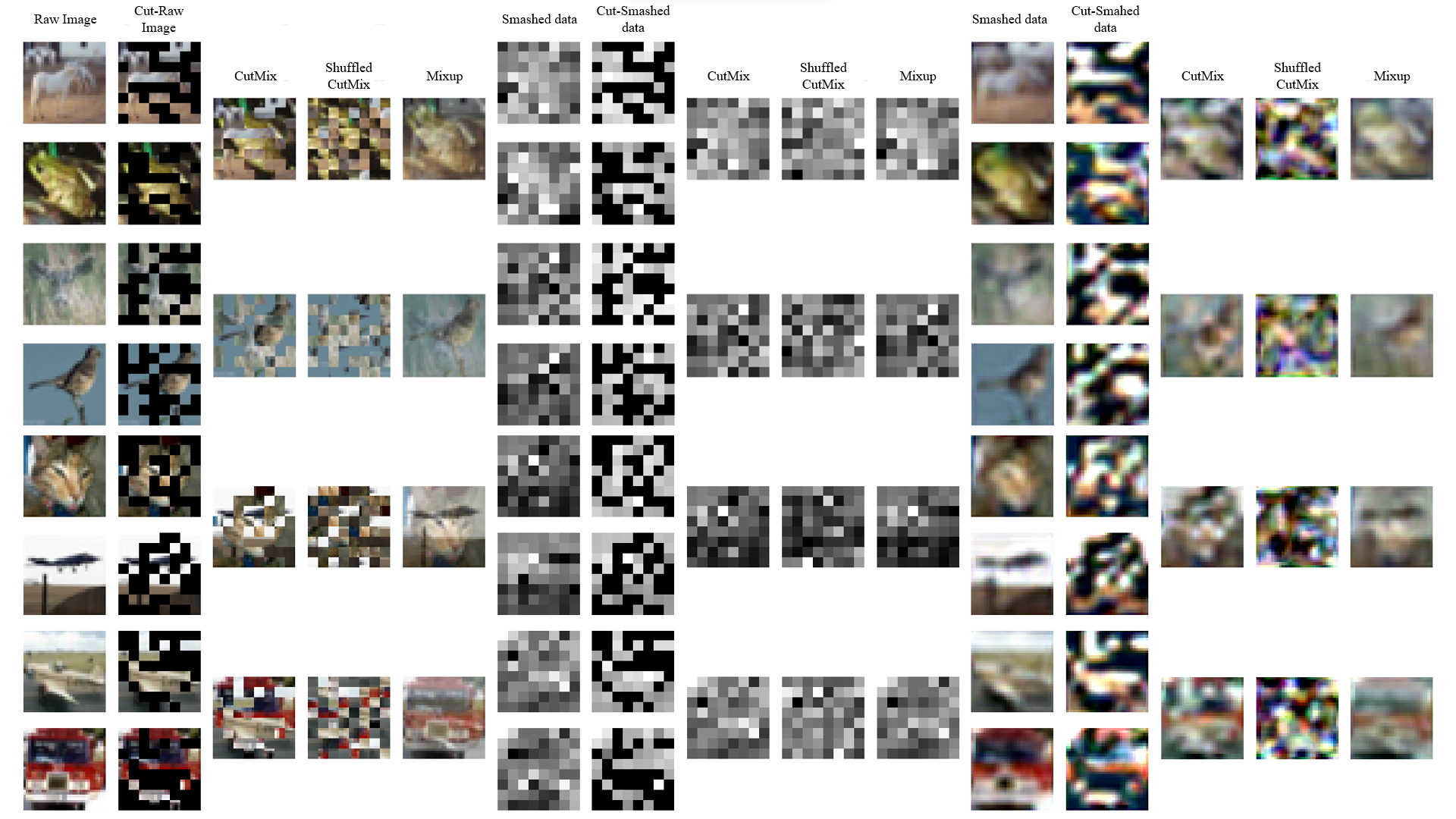}
    \centering
    \caption{Raw images.}
    \label{fig:examp1_apen}
    \end{subfigure}
    \hfill
   \begin{subfigure}[ht]{0.33\textwidth}
    \includegraphics[trim={13.6cm 0 13.6cm 0cm},clip,width=\textwidth]{images/vis_appen.png}
    \centering
    \caption{Smashed data.}
    \label{fig:examp2_apen}
    \end{subfigure}
    \hfill
   \begin{subfigure}[ht]{0.33\textwidth}
    \includegraphics[trim={27cm 0cm 0 0cm},clip,width=\textwidth]{images/vis_appen.png}
    \centering
    \caption{Reconstruction from smashed data.}
    \label{fig:examp3_apen}
    \end{subfigure}
    \caption{Additional examples of data with different operations at raw images, smashed data, and reconstructed images.}
    \label{cccc}
\end{figure*}

Additional examples of mix operations on raw images, smashed data and its corresponding reconstruction images are shown in the next page. All the images shown in \Cref{cccc} are based on mixtures of two images (2-way). 
\end{document}